\documentclass[journal]{IEEEtran}

\usepackage{float}
\usepackage{graphicx}
\usepackage{amsmath}
\usepackage{amssymb}
\usepackage{multirow}
\usepackage{booktabs}
\usepackage{wrapfig}
\usepackage{amsmath}

\graphicspath{{./images/}}
\DeclareGraphicsExtensions{.png , .jpg, .jpeg, .eps}

\usepackage{hyperref}
\usepackage{comment}
\usepackage[table,xcdraw]{xcolor}

\usepackage{multicol}

\usepackage[noadjust]{cite}

\definecolor{awesome}{rgb}{1.0, 0.13, 0.32}
\definecolor{blue-violet}{rgb}{0.54, 0.17, 0.89}

\newcommand{\ie}{{\em i.e.}}

\newcommand{\eg}{{\em e.g.}}
\newcommand{\etal}{{\em et al.}}

\newcommand{\Fig}[1]{Fig. \ref{fig:#1}}

\newcommand{\Sect}[1]{Sect. \ref{sec:#1}}
\newcommand{\SSect}[1]{Sect. \ref{ssec:#1}}

\newcommand{\pdata}{\mathbf{p}}
\newcommand{\vstate}{\mathbf{m}}
\newcommand{\action}{\mathbf{a}}
\newcommand{\navcom}{c}
\newcommand{\PF}{P}
\newcommand{\MF}{M}
\newcommand{\PMF}{J}
\newcommand{\actionbranch}{A}
\newcommand{\driver}{F}
\newcommand{\weights}{\mathbf{w}}

\begin{document}

\title{Multimodal End-to-End Autonomous Driving}

\author{Yi Xiao, Felipe Codevilla, Akhil Gurram, Onay Urfalioglu, Antonio M. L\'opez
\thanks{Yi, Felipe, and Antonio are with the Computer Vision Center (CVC) and the Univ. Aut\`onoma de Barcelona (UAB). Akhil and Onay are with Huawei GRC in Munich. CVC members acknowledge the financial support by the Spanish TIN2017-88709-R (MINECO/AEI/FEDER, UE). Antonio M. L\'opez acknowledges the financial support by ICREA under the ICREA Academia Program. Felipe Codevilla acknowledges Catalan AGAUR for his FI grant 2017FI-B1-00162. Yi Xiao acknowledges the Chinese Scholarship Council (CSC), grant number 201808390010. We also thank the Generalitat de Catalunya CERCA Program, as well as its ACCIO agency.}}

\maketitle

\begin{abstract}

A crucial component of an autonomous vehicle (AV) is the artificial intelligence (AI) is able to drive towards a desired destination. Today, there are different paradigms addressing the development of AI drivers. On the one hand, we find modular pipelines, which divide the driving task into sub-tasks such as perception and maneuver planning and control. On the other hand, we find end-to-end driving approaches that try to learn a direct mapping from input raw sensor data to vehicle control signals. The later are relatively less studied, but are gaining popularity since they are less demanding in terms of sensor data annotation. This paper focuses on end-to-end autonomous driving. So far, most proposals relying on this paradigm assume RGB images as input sensor data. However, AVs will not be equipped only with cameras, but also with active sensors providing accurate depth information ({\eg}, LiDARs). Accordingly, this paper analyses whether combining RGB and depth modalities, {\ie} using RGBD data, produces better end-to-end AI drivers than relying on a single modality. We consider multimodality based on early, mid and late fusion schemes, both in multisensory and single-sensor (monocular depth estimation) settings. Using the CARLA simulator and conditional imitation learning (CIL), we show how, indeed, early fusion multimodality outperforms single-modality. 

\end{abstract}


\IEEEpeerreviewmaketitle

%
\section{Introduction}
\IEEEPARstart{A}{utonomous} vehicles (AVs) are core for future mobility. Thus, it is essential to develop artificial intelligence (AI) for driving AVs. Two main paradigms are under research, namely, \emph{modular pipelines} and \emph{end-to-end driving}. 

The modular paradigm attaches to the traditional divide-and-conquer engineering principle, since AI drivers rely on modules with identifiable responsibilities; for instance, to provide environmental perception \cite{Franke:2017, Janai:2017}, as well as route planning and maneuver control \cite{Paden:2016, Schwarting:2018}. Perception itself is already especially complex, since it involves sub-tasks such as object detection \cite{Ren:2015, Yang:2018, Qi:2018, Asvadi:2018, Pfeuffer:2018, Li:2019} and tracking \cite{Choi:2015, Xiang:2015, Wojke:2017, Bhat:2018, Dimitrievski:2019}, traffic sign recognition \cite{Zhu:2016}, semantic segmentation \cite{Long:2015, Noh:2015, Yu:2016, Uhrig:2016, Schneider:2017, Ha:2017}, monocular depth estimation \cite{Godard:2017, Gurram:2018, Gan:2018, Fu:2018}, SLAM and place recognition \cite{Bresson:2017, Tateno:2017, ChenZhang:2017, Shin:2018, Qiu:2018, Yin:2018, Zhu:2018}, etc. 

The end-to-end driving paradigm focuses on learning holistic models is able to directly map raw sensor data into control signals for maneuvering AVs \cite{Pomerleau:1989, LeCun:2005, Bojarski:2016, Xu:2017}, {\ie} without forcing explicit sub-tasks related to perception or planning. Thus, advocating for learning to perceive and act simultaneously, as humans do. Moreover, such sensorimotor models are obtained through a data-driven supervised learning process as is characteristic of modern AI. End-to-end driving models can accept high-level navigation commands \cite{Hubschneider:2017, Codevilla:2018, Wang:2018, Liang:2018}, or be restricted to specific navigation sub-tasks such as lane keeping \cite{Innocenti:2017, Chen:2017, Eraqi:2017} and longitudinal control \cite{George:2018}.

Driving paradigms are highly relying on convolutional neural netwoks (CNNs). In this context, one of the main advantages of modular pipelines is the ability to explain the decisions of the AI driver in terms of its modules; which is more difficult for pure end-to-end driving models \cite{Bojarski:2017, Kim:2017, LiMotoyoshi:2018}.  However, developing some of the critical modules of the modular paradigm requires hundreds of thousands of supervised data samples \cite{Sun:2017, Hestness:2017}, {\eg} raw sensor data with ground truth (GT). Since the GT is most of the times provided manually ({\eg} annotation of object bounding boxes \cite{Geiger:2012}, pixel-level delineation of semantic classes \cite{Cordts:2016}), this is an important bottleneck for this paradigm. Conversely, end-to-end approaches are able to learn CNN-based models for driving from raw sensor data ({\ie} without annotated GT) and associated supervision in terms of vehicle's variables ({\eg} steering angle, speed, geo-localization and orientation \cite{Santana:2016, Maddern:2017, Xu:2017}); note that such supervision does not require human intervention in terms of explicitly annotating the content of the raw sensor data. Moreover, end-to-end models are demonstrating an \emph{unreasonable} effectiveness in practice \cite{Bojarski:2016, Chen:2017, Codevilla:2018, George:2018}, which makes worth to go deeper in their study.

Although AVs will be multisensory platforms, equipping and maintaining on-board synchronized heterogeneous sensors is quite expensive nowadays. As a consequence, most end-to-end models for driving rely only on vision \cite{LeCun:2005, Chen:2015, Bojarski:2016, Xu:2017, Innocenti:2017, Chen:2017, Codevilla:2018, Sauer:2018, Muller:2018, YangZhang:2018, George:2018}, {\ie} they are visuomotor models. This is not bad in itself, after all, human drivers mainly rely on vision. However, multimodality has shown better performance in key perception sub-tasks such as object detection \cite{Gonzalez:2016, Gonzalez:2017, ChenMa:2017, Qi:2018, Asvadi:2018, Ku:2018, Pfeuffer:2018, Li:2019, Wu:2019}, tracking \cite{Dimitrievski:2019}, and semantic segmentation \cite{Schneider:2017, Ha:2017}. Thus, it is worth exploring multimodality for end-to-end driving.  

Accordingly, in this paper we address the question \emph{can an end-to-end driving model be improved by using multimodal sensor data over just relying on a single modality?} In particular, we assume color images (RGB) and depth (D) as single modalities, and RGBD as multimodal data. Due to its capability of accepting high level commands, this study is based on the CNN architecture known as \emph{conditional imitation learning} (CIL) \cite{Codevilla:2018}. We explore RGBD from the perspective of early, mid and late fusion of the RGB and D modalities. Moreover, as in many recent works on end-to-end driving \cite{Codevilla:2018, Muller:2018, Sauer:2018, Rhinehart:2018, Wang:2018,  Liang:2018, LiMotoyoshi:2018}, our experiments rely on the CARLA simulator \cite{Dosovitskiy:2017}. 

The presented results show that multimodal RGBD end-to-end driving models outperform their single-modal counterparts. Moreover, early fusion shows better performance than mid and late fusion. On the other hand, multisensory RGBD ({\ie} based on camera and LiDAR) still outperforms monocular RGBD; however, we conclude that it is worth pursuing this special case of single-sensor multimodal end-to-end models.     

We present the work as follows. \Sect{rw} reviews the related literature. \Sect{methods} presents the used CIL architecture from the point of view of early, mid, and late fusion. \Sect{experiments} summarizes the experimental setting and the obtained results. Finally, \Sect{conclusion} draws the main conclusions and future work.  

\section{Related work}
\label{sec:rw}
This section focuses on two main related topics: \emph{multimodal perception} and \emph{end-to-end driving models learned by imitation}.

\subsection{Multimodality}
Object detection is one of the perception tasks for which multimodality has received most attention. Enzweiler {\etal} \cite{Enzweiler:2011} developed a pedestrian detector using hand-crafted features and shallow classifiers combined as a mixture-of-experts (MoE), where multimodality relies on image luminance and stereo depth. Gonzalez {\etal} \cite{Gonzalez:2017} detected vehicles, pedestrians and cyclists---vulnerable road users (VRUs)---, using a multimodal MoE based on space-time calibrated RGB and LiDAR depth. Chen {\etal} \cite{ChenMa:2017} used calibrated RGB and LiDAR depth as input for a CNN-based detector of vehicles and VRUs, which is a current trend \cite{ChenMa:2017, Asvadi:2018, Pfeuffer:2018, Ku:2018, Wu:2019}. Some of these works are inspired by Faster R-CNN \cite{Ren:2015}, since they consist of a first stage for proposing regions potentially containing objects of interest, and a second stage performing the classification of those regions to provide final object detections; {\ie} following a mid-level (deep) fusion scheme where CNN layers of features from the different modalities are fused in both stages \cite{ChenMa:2017, Ku:2018, Wu:2019}. Other alternatives are early fusion at raw data level \cite{Pfeuffer:2018}, late fusion of independent detectors \cite{Asvadi:2018, Pfeuffer:2018}, or just using different modalities at separated steps of the detection pipeline \cite{Qi:2018}. Other approaches focus on multispectral appearance, as in Li {\etal} \cite{Li:2019}, where different fusion schemes for RGB and Far Infrared (FIR) calibrated images are compared. 

All these studies and recent surveys \cite{Arnold:2019, Feng:2019} show that detection accuracy increases with multimodality. Therefore, more perception tasks have been addressed under the multimodal approach. Dimitrievski {\etal}  \cite{Dimitrievski:2019} proposed a pedestrian tracker that fuses camera and LiDAR detections to solve the data association step of their tracking-by-detection approach. Schneider {\etal}  \cite{Schneider:2017} proposed a CNN architecture for semantic segmentation which performs a mid-level fusion of RGB and stereo depth, leading to a more accurate segmentation on small objects. Ha {\etal} \cite{Ha:2017} also proposed a mid-level RGB and FIR fusion approach in a CNN architecture for semantic segmentation. Piewak {\etal}  \cite{Piewak:2018} used a mid-level fusion of LiDAR and camera data to produce a Stixel representation of the driving scene, showing improved accuracy in terms of geometry and semantics of the resulting representation. 

In this paper, rather than focusing on individual perception tasks such as object detection, tracking or semantic segmentation, we challenge multimodality in the context of end-to-end driving, exploring early, mid and late fusion schemes.  

\subsection{End-to-end driving}
Pomerleau presented ALVINN three decades ago \cite{Pomerleau:1989}, a sensorimotor fully-connected shallow neural network that was able to perform end-to-end road following assuming no obstacles. ALVINN controlled a CMU's van, NAVLAB, along a $400\mbox{m}$ straight path, at ${\sim2}$ {Km}/h and under good weather conditions. Although the addressed scenario is extremely simple compared to driving in real traffic, it was already necessary to simulate data for training the sensorimotor model and, in fact, camera images ($30\times32$ pixels, blue channel) were already combined by early fusion with laser range finder data ($8\times32$ depth cells). 
LeCun {\etal} \cite{LeCun:2005} trained end-to-end a 6-layer CNN for off-road obstacle avoidance using image pairs (from a stereo rig) as input. Such CNN was able to control a $50\mbox{cm-length}$ four-wheel truck, DAVE, for avoiding obstacles at a speed of ${\sim7}$ {Km}/h. During data collection for training, the truck was remotely controlled by a human operator, thus, the CNN was trained according to imitation learning in our terminology (or teleoperation-based demonstration \cite{Argall:2009}). More recently, Bojarski {\etal} \cite{Bojarski:2016} developed a vision-based end-to-end driving CNN which was able to control the steering wheel of a real car in different traffic conditions. 
Still, lane and road changing are not considered, neither stop-and-go maneuvers since throttle and brake are not controlled. 

These pioneering works inspired new proposals based on imitation learning for CNNs. Eraqi {\etal} \cite{Eraqi:2017} applied vision-based end-to-end control of the steering angle (neither throttle nor break), focusing on including temporal reasoning by means of long short-term memory recurrent neural networks (LSTMs). Training and testing were done in the Comma.ai dataset \cite{Santana:2016}. George {\etal} \cite{George:2018} applied similar ideas for controlling the speed of the car. Xu {\etal} \cite{Xu:2017} presented the BDD dataset and focused on vision-based prediction of the steering angle using a fully convolutional network (FCN) and a LSTM, forcing semantic segmentation as auxiliary training task. Innocenti {\etal} \cite{Innocenti:2017} performed vision-based end-to-end steering angle prediction for lane keeping on private datasets, and Chen {\etal} \cite{Chen:2017} in the Comma.ai dataset.


Affordances have been proposed as intermediate tasks between enviromental perception and prediction of the vehicle control parameters \cite{Chen:2015, Sauer:2018}. Affordances do not require to solve perception sub-tasks such as explicit object detection, etc; but they form a compact set of factors that influence driving according to prior human knowledge. Chen {\etal} \cite{Chen:2015} evaluated them on the TORCS simulator \cite{Wymann:2014}, so in car racing conditions (no pedestrians, no intersections, etc.) under clean and dry weather; while Sauer {\etal} \cite{Sauer:2018} used the CARLA simulator, which supports regular traffic conditions under different lighting and weather \cite{Dosovitskiy:2017}. Muller {\etal} \cite{Muller:2018} developed a vision-based CNN with an intermediate road segmentation task for learning to perform vehicle maneuvers in a semantic space; the driving policy consists of predicting waypoints within the segmented road and applying a low-level PID controller afterwards. Training and testing are done in CARLA, but neither incorporating other vehicles nor pedestrians. Using LiDAR data, Rhinehart {\etal} \cite{Rhinehart:2018} combined imitation learning and model-based reinforcement learning to predict expert-like vehicle trajectories, relying on CARLA but without dynamic traffic participants.  

These end-to-end driving models do not accept high-level navigation instructions such as \emph{turn left at the next intersection} (without providing explicit distance information), which can come from a global planner or just as voice commands from a passenger of the AV. Hubschneider {\etal} \cite{Hubschneider:2017} proposed to feed a turn indicator in the vision-based CNN driving model by concatenating it with features of a mid-level fully connected layer of the CNN. Codevilla {\etal} \cite{Codevilla:2018} proposed a more effective method, in which a vision-based CNN consisting of an initial block agnostic to particular navigation instructions, and a second block branched according to a subset of navigation instructions (at next intersection turn-left/turn-right/go-straight, or just keep lane). In the first block, vehicle information is also incorporated as mid-level feature of the CNN; in particular, current speed is used since the CNN controls the steering angle, throttle, and break (Yang {\etal} \cite{YangZhang:2018} also reported the usefulness of speed feedback in end-to-end driving). Experiments are performed in CARLA for different traffic situations (including other vehicles and pedestrians), lighting and weather conditions. The overall approach is termed as conditional imitation learning (CIL). In fact, Muller {\etal} and Sauer {\etal} leveraged from CIL. Liang {\etal} \cite{Liang:2018} also used CIL as imitation learning stage before refining the resulting model by applying reinforcement learning. Wang {\etal} \cite{Wang:2018} used CIL too, but incorporating ego-vehicle heading information at the same CNN-layer level as speed.

These works focus on vision-based end-to-end driving. Here, we explore muldimodal end-to-end driving based on RGB and depth; which can be complementary to most of the cited papers. Without losing generality, we chose CIL as core CNN architecture due to its effectiveness and increasing use.

Focusing on multimodality, Sobh {\etal} \cite{Sobh:2018} used CARLA to propose a CIL-based driving approach modified to process camera and LiDAR data. In this case, the information fusion is done by a mid-level approach; in particular, before fusion, RGB images are used to generate a semantic segmentation which corresponds to one of the information streams reaching the fusion layers, and there are two more independent streams based on LiDAR, one encoding a bird view and the other a polar grid mapping. Khan {\etal} \cite{Khan:2019} also used CARLA to propose an end-to-end driving CNN based on RGB and depth images, which predicts only the steering angle, assuming that neither other vehicles nor pedestrians are present. In a first step, the CNN is trained only using depth information (taken as the Z-buffer produced by UE4,
the game engine behind CARLA). This CNN has an initial block of layers (CNN encoder) that outputs depth-based features, which are later used to predict the steering angle with a second block of fully connected layers. In a second step, this angle-prediction block is discarded and replaced by a new fully connected one. This new block relies on the fusion of the depth-based features and a semantic segmentation produced by a new CNN block that processes the RGB image paired with the depth image. During training, semantic segmentation is conditioned to depth-based features due to the fusion block and back-propagation. This approach can be considered a type of mid-level fusion.

In contrast to these multimodal end-to-end driving approaches, we assess early, mid and late level fusion schemes without forcing intermediate representations which are not trivial to obtain ({\eg} semantic segmentation is an open problem in itself). Moreover, we run CARLA 
benchmark \cite{Dosovitskiy:2017}, which includes dynamic actors (vehicles and pedestrians) and generalization conditions (unseen town and weather). We show that CIL and early fusion produce state-of-the-art results. 
\begin{figure*}[t!]
    \centering
    \includegraphics[width=0.9\textwidth]{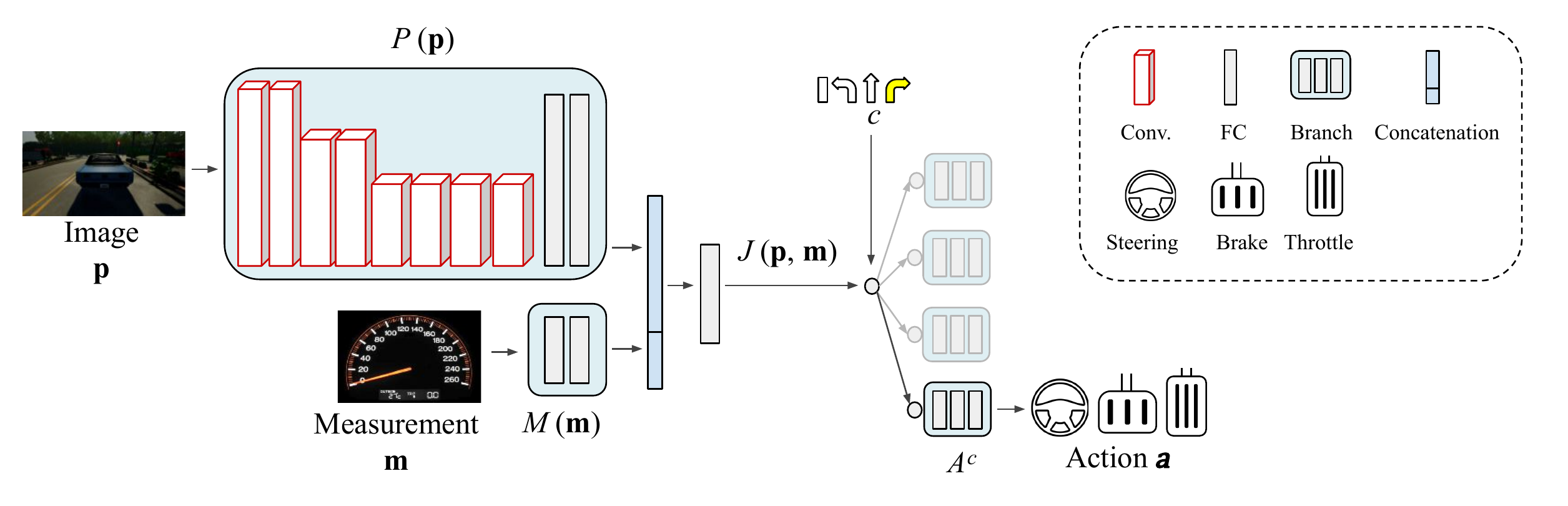}
    \caption{CIL branched architecture: vehicle maneuvers (actions) in the form of the triplet $<\!\!\!\mbox{steering angle, throttle, brake}\!\!\!>$, depend on a high-level route navigation command (branch selector) running on $\{\mbox{turn-left}, \mbox{turn-right}, \mbox{go-straight}, \mbox{continue}\}$, as well as on the world observations in the form of perception data ({\eg} a RBG image) and vehicle state measurements ({\eg} speed).}
    \label{fig:CIL}
\end{figure*} 

\section{Multimodal Fusion}
\label{sec:methods}
We first detail CIL \cite{Codevilla:2018}, and then show how we adapt it to leverage from multimodal perception data.  

\subsection{Base CIL architecture}

\Fig{CIL} shows the CNN implementing CIL. The observations (CIL's input) are twofold, perception data, $\pdata$, and vehicle's state measurements, $\vstate$. The action (CIL's output), $\action$, consists of vehicle controls for maneuvering. CIL includes a CNN block to extract perception features, $\PF(\pdata)$; and a block of fully connected layers to extract measurement features $\MF(\vstate)$. A joint layer of features is formed by appending $\PF(\pdata)$ and $\MF(\vstate)$; which is further processed by a new fully connected layer to obtain the joint features $\PMF(<\!\!\PF(\pdata),\MF(\vstate)\!\!>)$, or just $\PMF(\pdata,\vstate)$ simplifying the notation. Up to this point of the neural network, the processing done with the observations is common to any driving maneuver/action. However, many times, the autonomous vehicle reaches ambiguous situations which require to incorporate informed decisions. For instance, when reaching a cross intersection, without incorporating a route navigation command ({\eg} from a global trajectory plan), the vehicle could only take a random decision about turning or going straight. Thus, the end-to-end driving CNN must incorporate high-level commands, $\navcom$, such as \emph{`in the next intersection turn left'}, or \emph{`turn right'}, or  \emph{`go straight'}. Moreover, $\action$ will take very different values depending on $\navcom$. Thus, provided $\navcom$ takes discrete values, having specialized neural network layers for each maneuver can be more accurate a priori. All this is achieved in the CIL proposal by incorporating fully connected maneuver/action branches, $\actionbranch^\navcom$, selected by $\navcom$ (both during CNN training and vehicle self-driving). 


We follow the CIL architecture proposed in \cite{Codevilla:2018}. Therefore, $\pdata$ is a RGB image of $200\times88$ pixels and 8 bits at each color channel, $\vstate$ is a real value with the current speed of the vehicle, and $\action$ consists of three real-valued signals which set the next maneuver in terms of steering angle, throttle, and brake. Thus, the idea is to perform vision-based self-driving, as well as taking into account the vehicle speed to apply higher/lower throttle and brake for the same perceived traffic situation. In \cite{Codevilla:2018}, the focus is on handling intersections, then the considered $\navcom$ values are $\{\mbox{\emph{turn-left, turn-right, go-straight, continue}}\}$, where the last refers to just keep driving in the current lane and the others inform about what to do when reaching next intersection (which is an event detected by the own CNN). Accordingly, there are four branches $\actionbranch^\navcom$. Therefore, if we term by $\driver$ the end-to-end driver, we have $\driver(\pdata,\vstate,\navcom)=\actionbranch^\navcom(\PMF(\pdata,\vstate))$. As shown in \cite{Codevilla:2018}, this manner of explicitly taking into account high-level navigation commands is more effective than other alternatives.  


\subsection{Fusion schemes}

\begin{figure*}[t!]
    \centering
    \includegraphics[width=0.9\textwidth]{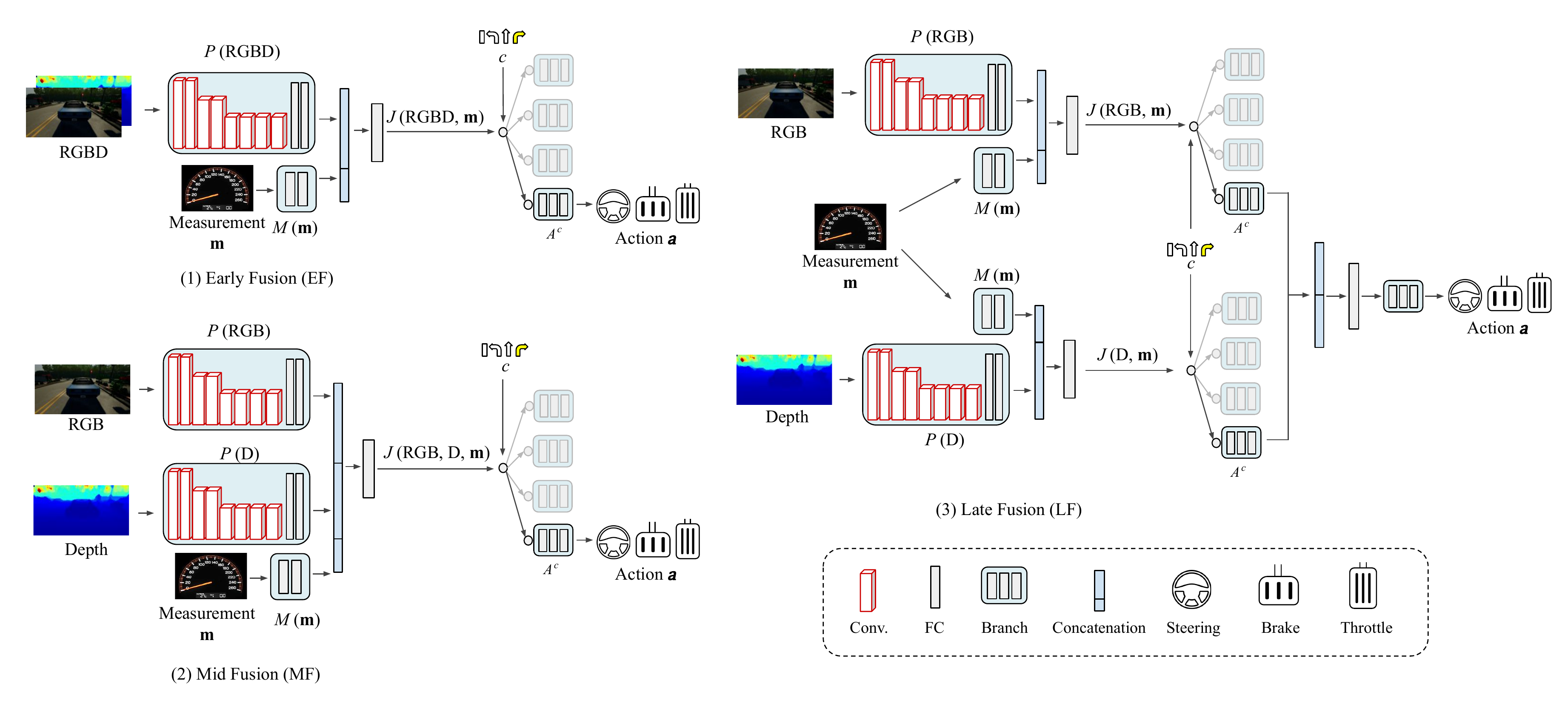}
    \caption{Network Architectures - we explore RGBD from the perspective of early, mid and late fusion of the RGB and Depth (D) modalities. (1) Early Fusion: the raw RGB and D channels are the input of the CIL architecture; (2) Mid Fusion: intermediate CIL feature layers  from  RGB and D streams are fused; (3) Late Fusion: the output (maneuver controls) of the RGB and D CIL streams are fused to output the final values after further neural processing.}
    \label{fig:FusionArchitecture}
\end{figure*}

\Fig{FusionArchitecture} illustrates how we fuse RGB and depth information following mid, early and late fusion approaches.

\emph{Early fusion:} with respect to the original CIL we only change the number of channels of $\pdata$ from three (RGB) to four (RGBD). The CIL network only changes the first convolutional layer of $\PF(\pdata)$ to accommodate for the extra input channel, the rest of the network is equal to the original. 

\emph{Mid fusion:}we replicate twice the perception processing $\PF(\pdata)$. One of the $\PF(\pdata)$ blocks processes only RGB images, the other one only depth images. Then, we build the joint feature vector $<\!\!\PF(\mbox{RGB}),\PF(\mbox{D}),\MF(\vstate)\!\!>$ which is further processed to obtain $\PMF(\mbox{RGB},\mbox{D},\vstate)$. From this point, the branched part of CIL is the same as in the original architecture. 

\emph{Late fusion:} we replicate twice the full CIL architecture. Thus, RGB and depth channels are processed separately, but the measurements are shared as input. Hence, we run  $\actionbranch^\navcom(\PMF(\mbox{RGB},\vstate))$ and $\actionbranch^\navcom(\PMF(\mbox{D},\vstate))$, and their outputs are concatenated and further processed by a module of fully connected layers, the output of which conveys the final action values. Note that this is a kind of mixture-of-experts approach, where the two experts are jointly trained. 

As is common practice in the literature, we assume a pixel-level correspondence of all channels and normalize all of them to be in the same magnitude range (we normalize depth values to match the range of color channels, {\ie} $[0\ldots255]$). 

\subsection{Loss function}
Given a predicted action $\action$, its ground truth $\action^{gt}$, and a vector of weights $\weights$, we use the L1 loss $\ell_{act}(\action, \action^{gt}, \weights) = \sum_i^n \arrowvert w_i(a_i-a^{gt}_i) \arrowvert$, with $n=3$ (steering angle, throttle, brake). Note that when computing $\action$, only one $\actionbranch^\navcom$ branch is active at a time. In particular, the one selected by the particular command $\navcom$ associated to the current input data ($\pdata, \vstate$). We make this fact explicit by changing the notation to $\ell_{act}(\action, \action^{gt}, \weights; \navcom)$. 

In addition, as in other computer vision problems addressed by deep learning \cite{Doersch:2017, Kendall:2018}, we empirically found that using multi-task learning helps to obtain more accurate CIL networks. In particular, we add an additional branch of three fully connected layers to predict current vehicle speed from the perception data features $\PF(\pdata)$.
This prediction relies on a L1 loss $\ell_{sp}(s, s^{gt})=\arrowvert s-s^{gt} \arrowvert$, where $s$ is the predicted speed and $s^{gt}$ is the ground truth speed which, in this case, is already available since it corresponds to the measurement used as input. Speed prediction is only used during training. 

Thus, all networks, {\ie} both single- and multimodal, are trained according to the same total loss  $\ell(\action, \action^{gt}, \weights; \navcom; s, s^{gt})=\beta\ell_{act}(\action, \action^{gt}, \weights; \navcom)+(1-\beta)\ell_{sp}(s, s^{gt})$, where $\beta$ is used to balance the relevance of $\ell_{act}$ and $\ell_{sp}$ losses. 

\section{Experiments}
\label{sec:experiments}

We start by summarizing the environment we use for our experiments, {\ie} CARLA (\SSect{carla-environment}). Next, we describe the driving benchmark available in CARLA (\SSect{carla-benchmarks}), the dataset we use for training our AI drivers (\SSect{carla-dataset}), and the training protocol that we follow (\SSect{carla-training}). Finally, we present and discuss the obtained results (\SSect{carla-results}).

\subsection{Environment}
\label{ssec:carla-environment}

In order to conduct our experiments, we rely on the open source driving simulator CARLA \cite{Dosovitskiy:2017}. There are several reasons for this. First, many recent previous works on end-to-end driving  rely on CARLA \cite{Codevilla:2018, Muller:2018, Sauer:2018, Rhinehart:2018, Wang:2018,  Liang:2018, LiMotoyoshi:2018}; thus, we can compare our results with the previous literature. Second, it seems that for some scenarios the end-to-end paradigm may need exponentially more training samples than the modular one \cite{Shalev:2016}, so there is a trade-off between collecting driving runs (for the end-to-end paradigm) and manually annotating on-board acquired data (for the modular paradigm) which, together with the gigantic effort needed to demonstrate that an AV outperforms human drivers, really encourages to rely on simulators during the development of AI drivers \cite{Kalra:2016}. Yet, a third and core reason is specific for end-to-end driving models. In particular, in \cite{Codevilla:2018b} it is demonstrated that current offline evaluation metrics ({\ie} based on static datasets) for assessing end-to-end driving models do not correlate well-enough with actual driving, an observation also seen in \cite{Bewley:2018}; therefore, it is really important to evaluate these driving models in an on-board driving regime, which is possible in a realistic simulator such as CARLA.  

\begin{figure}
\centering
\includegraphics[height=2.5cm]{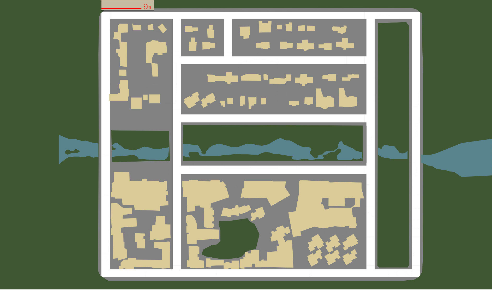}
\includegraphics[height=2.5cm]{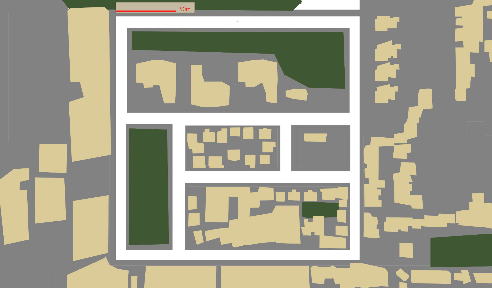}
\caption{Bird-view road maps of Town 1 (left) and Town 2 (right).}
\label{fig:towns}
\end{figure}

\begin{figure}
\centering
\includegraphics[width=0.49\columnwidth]{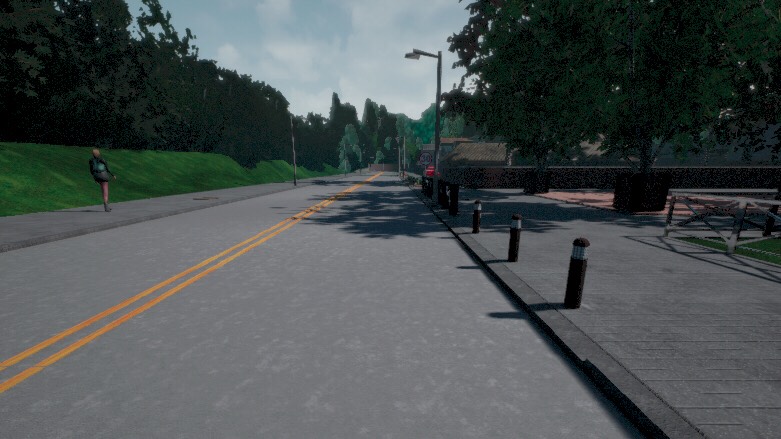}\hspace{0.05cm}\includegraphics[width=0.49\columnwidth]{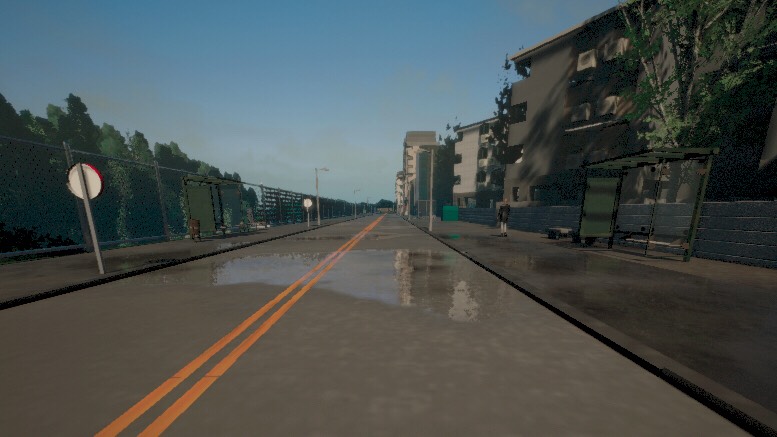}\vspace{0.05cm}

\includegraphics[width=0.49\columnwidth]{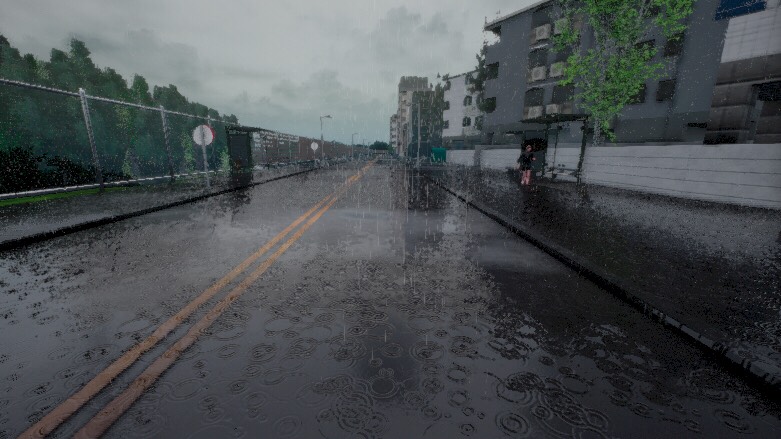}\hspace{0.05cm}\includegraphics[width=0.49\columnwidth]{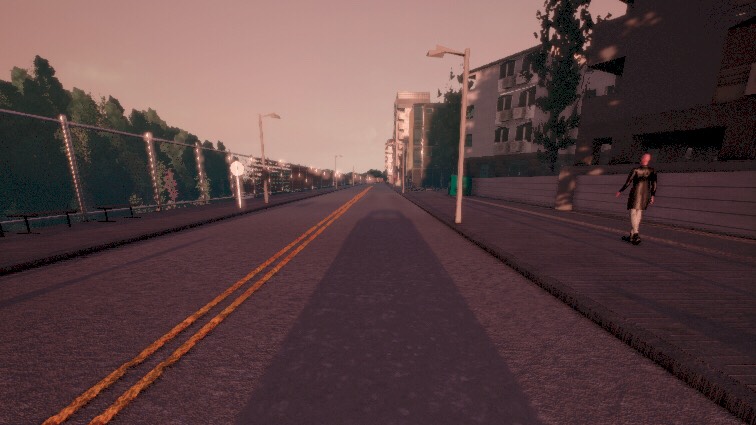}\vspace{0.05cm}

\includegraphics[width=0.49\columnwidth]{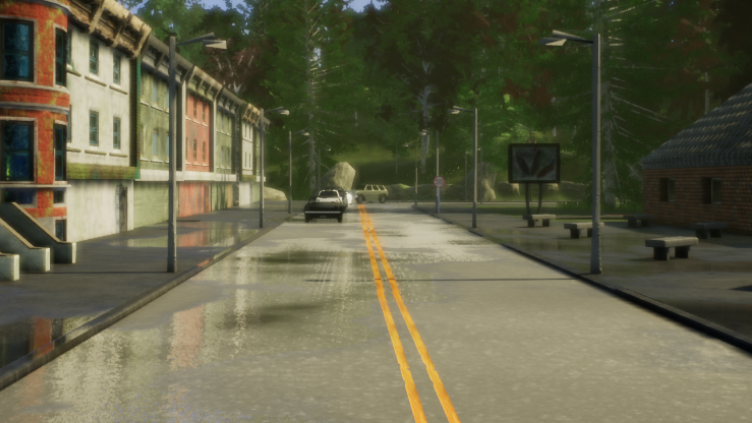}\hspace{0.05cm}\includegraphics[width=0.49\columnwidth]{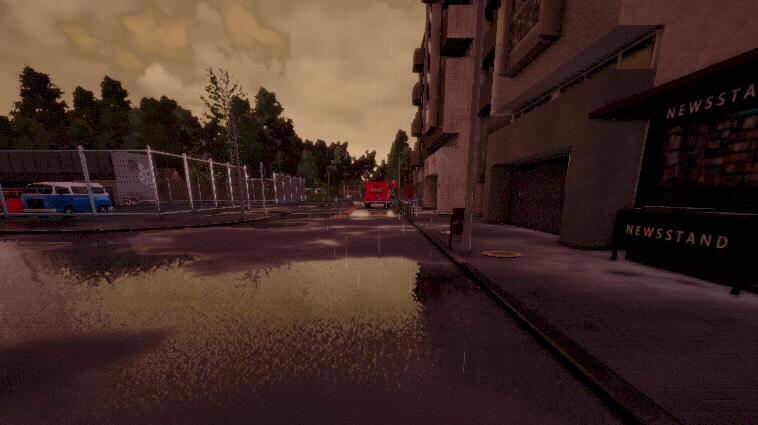}
\caption{Top, from Town 1: clear noon (left) and clear after rain (right). Middle, from Town 1: heavy rain noon (left) and clear sunset (right). Bottom, from Town 2: wet cloudy noon (left) and soft rainy sunset (right).}
\label{fig:weathers}
\end{figure}

Briefly, CARLA contains two towns (\Fig{towns}) based on two-directional roads with turns and intersections, buildings, vegetation, urban furniture, traffic signs, traffic lights, and dynamic objects such as vehicles and pedestrians. Town~1 deploys 2.9Km of road and 11 intersections, while Town~2 contains 1.4Km of road and 8 intersections. The different towns can be travelled under six different weather conditions (\Fig{weathers}): `clear noon', `clear after rain', `heavy rain noon', and `clear sunset', `wet cloudy noon' and `soft rainy sunset'.

\subsection{Driving benchmark}
\label{ssec:carla-benchmarks}

CARLA was deployed with a benchmarking infrastructure for assessing the performance of AI drivers \cite{Dosovitskiy:2017}. Thus, it has been used in the related literature since then and we follow it here too. Four increasingly difficult \emph{driving tasks} are defined: 
\begin{itemize}
\item \emph{straight}: the destination point is straight ahead from the starting point but no dynamic objects are present; 
\item \emph{one turn}: destination is one turn away from the starting point, no dynamic objects; 
\item \emph{navigation}: no restriction on the location of the destination and starting points, no dynamic objects; 
\item \emph{navigation with dynamic obstacles}. 
\end{itemize}

\begin{figure*}[t!]
\centering
\includegraphics[width=0.38\columnwidth]{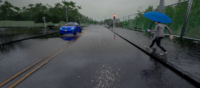}\hspace{0.05cm}\includegraphics[width=0.38\columnwidth]{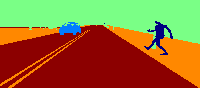}\hspace{0.05cm}\includegraphics[width=0.38\columnwidth]{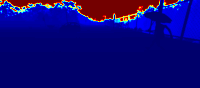}\hspace{0.05cm}\includegraphics[width=0.38\columnwidth]{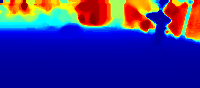}\hspace{0.05cm}\includegraphics[width=0.38\columnwidth]{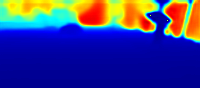}\hspace{0.20cm}\includegraphics[width=0.05\columnwidth]{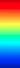}
\caption{From left to right: original RGB image, semantic segmentation ground truth (for the five considered classes), CARLA depth ground truth, post-processed to be closer to the capabilities of an \emph{active} depth Sensor, and monocular depth estimation from a model trained using such a depth.}
\label{fig:rgb-gt-monocular}
\end{figure*}

For each driving task, an AI driver is assessed over a total of $E_T$ driving episodes. Each episode has its own starting and destination points with an associated topological route. An episode is considered as successful if the AI driver completes the route within a time budget. Collisions do not lead to the termination of an episode unless the AV runs in time-out as a consequence. If we term as $E_S$ the total number of successfully completed episodes by the assessed AI driver, then its \emph{success rate} is defined as $100\times(E_S/E_T)$. $E_T$ is determined by the selected town and weather conditions. 

Table \ref{tab:t-v-t} shows how the benchmark organizes towns and weather conditions for training, validation, and testing; where, irrespective of the town and weather, validation and testing is always based on episodes, not in pre-recorded datasets, while training requires pre-recording a dataset. Validation is performed to select a driving model among those trained as different trials from the same training dataset, while testing is performed for actually benchmarking the selected models. 

Regarding town and weather conditions, the benchmark establishes four main town-weather blocks under which the four driving tasks must be tested, assuming 25 episodes for each considered weather. Therefore, for each block, the $E_T$ value is different as we can deduce from Table \ref{tab:t-v-t}. In particular, these are the town-weather blocks defined in the benchmark with their respective $E_T$ value:  

\emph{Training conditions}: driving ({\ie} running the episodes) in the same conditions as the training set (Town 1, four weather conditions), thus, $E_T=100$;

\emph{New town}: driving under the four weather conditions of the training set but in Town 2, $E_T=100$;

\emph{New weather}: driving in Town 1 but under the two weather conditions not seen at training time, $E_T=50$;

\emph{New town \& weather}: driving in conditions not included in the training set (Town 2, two weather conditions), $E_T=50$. 

\begin{table}
\centering
\caption{Training, validation, and testing settings. Training is based on a pre-recorded dataset. Validation and testing are based on actual driving episodes. Grey means `not used'.}
\label{tab:t-v-t}
\resizebox{0.7\columnwidth}{!}{%
\begin{tiny}
\begin{tabular}{l|c|c|c|}
\cline{2-4}
                                        & Training                                         & Validation                                                                & Testing \\ 
                                        & (dataset)                                        & (episodes)                                                                & (episodes) \\ \hline
\multicolumn{1}{|l|}{Wet cloudy noon}   & \cellcolor[HTML]{EFEFEF}                         & \cellcolor[HTML]{EFEFEF}                                                  & \\ \cline{1-1}
\multicolumn{1}{|l|}{Soft rainy sunset} & \multirow{-2}{*}{\cellcolor[HTML]{EFEFEF}}       &                                                                           &  \\ \cline{1-1}
\multicolumn{1}{|l|}{Clear noon}        &                                                  & \multirow{-2}{*}{\begin{tabular}[c]{@{}c@{}}Towns \\ 1 \& 2\end{tabular}} & \\ \cline{1-1}
\multicolumn{1}{|l|}{Clear after rain}  &                                                  & \cellcolor[HTML]{EFEFEF}                                                  & \\ \cline{1-1}
\multicolumn{1}{|l|}{Clear sunset}      &                                                  & \cellcolor[HTML]{EFEFEF}                                                  & \\ \cline{1-1}
\multicolumn{1}{|l|}{Heavy rain noon}   & \multirow{-4}{*}{\cellcolor[HTML]{FFFFFF}Town 1} & \multirow{-3}{*}{\cellcolor[HTML]{EFEFEF}}                                & \multirow{-6}{*}{\begin{tabular}[c]{@{}c@{}}Towns\\  1 \& 2\end{tabular}} \\ \hline
\end{tabular}%
\end{tiny}
}
\end{table}



\subsection{Training dataset}
\label{ssec:carla-dataset}

In order to train our CNNs, we use the same dataset as in \cite{Codevilla:2018b} since it corresponds to 25h of driving in Town~1, balancing weather conditions (Table \ref{tab:t-v-t}). Briefly, this dataset was collected by a hard-coded auto-pilot with access to all the privileged information of CARLA required for driving like an expert. The auto-pilot kept a constant speed of 35 km/h when driving straight and reduced the speed when making turns. 
Images were recorded at 10fps from three cameras: a central forward-facing one and two lateral cameras facing $30^\circ$ left and right. The central camera is the only one used for self-driving, while the images coming from the lateral cameras are used only at training time to simulate episodes of recovering from driving errors as can be done with real cars \cite{Bojarski:2016} (the protocol for injecting noise follows \cite{Codevilla:2018}). Overall, the dataset contains ${\sim2.5}$ millions of RGB images of $800\times600$ pixels resolution, with associated ground truth (see \Fig{rgb-gt-monocular}) consisting of corresponding images of dense depth and pixel-wise semantic classes (semantic segmentation), as well as meta-information consisting of the high-level commands provided by the navigation system (continue in lane, at next intersection go straight, turn left or turn right), and car information such as speed, steering angle, throttle, and braking. In this work, we use perfect semantic segmentation to develop an upper-bound driver. Since we focus on end-to-end driving, the twelve semantic classes of CARLA are mapped to five which we consider sufficient to develop such an upper-bound. In particular, we keep the original \emph{road-surface}, \emph{vehicle}, and \emph{pedestrian}, while \emph{lane-marking} and \emph{sidewalk} are mapped as \emph{lane-limits} (Town~1 and Town~2 only have roads with one go and one return lane, separated by double continuous lines), and the remaining seven classes are mapped as \emph{other}. 

Focusing on depth information, as is common in the literature, we assume that RGB images have associated dense depth information; for instance, Premebida {\etal} \cite{Premebida:2014} obtained it from LiDAR point clouds. In CARLA, the depth ground truth is extremely accurate since it comes directly from the Z-buffer used during simulation rendering. In particular, depth values run from 0 to 1,000 meters and are codified with 24 bits, which means that depth precision is of ${\sim1/20}$ {mm}. This distance range coverage and depth precision is far beyond from what even \emph{active} sensors can provide. Therefore, we post-process depth data to make it more realistic. In particular, we take as a realistic sensor reference the Velodyne information of KITTI dataset \cite{Geiger:2012}. First we trim depth values to consider only those within the 1 to 100 meters interval, {\ie} pixels of the depth image with values outside this range are considered as not having depth information. Second, we re-quantify the depth values to have an accuracy of ${\sim4}$ {cm}. Third, we perform inpainting to fill-in the pixels with no information. Finally, we apply a median filter to avoid having perfect depth boundaries between objects. The new depth images are used both during training and testing. \Fig{rgb-gt-monocular} shows an example of a depth image from CARLA and its corresponding post-processed version. 

\begin{table}
\caption{$V_P$ for five training runs for RGB only, Depth (D) only, and RGBD combined by early (EF), mid (MF), or late (LF) fusion. Depth: from an active sensor or estimated from RGB images.}\label{tab:VV}
\centering
\resizebox{0.25\textwidth}{!}{
\begin{tabular}{@{}rcccccccc@{}}
\toprule
& RGB  & \multicolumn{4}{c}{Active} & \multicolumn{2}{c}{Estimation}\\
& & D & EF & MF & LF & D & EF \\
\midrule
1  & \textbf{48}  & \textbf{74}  & \textbf{91} & 61 & 60 & 51 & 42 \\
2  & 36  & 67 & 71 & 71 & 63 & 49 & 44 \\
3  & 46  & 73 & 75 & 58 & \textbf{67} & 46 & \textbf{51} \\
4  & 40  & 68 & 71 & \textbf{74} & 60 & \textbf{59} & 46 \\
5  & 36  & 68 & 77 & 52 & 62 & 51 & 49 \\
\bottomrule
\end{tabular}
}
\end{table}

\begin{table*}[!t]
\centering
\caption{Mean and standard deviation of success rates on the original CARLA Benchmark, by running it three times. CIL based on perfect semantic segmentation (SS) acts as upper bound. Excluding SS, for models tested under the same  environment and traffic conditions, we show in bold the higher means and we underline similar success rates considering standard deviations too. }\label{tab:CARLA-benchmark}
\resizebox{1.0\linewidth}{!}{
\begin{tabular}{@{}lccccccccccccccccc@{}}
\bottomrule
&&&\multicolumn{4}{c}{Active} & \multicolumn{2}{c}{Estimated}&&&& \multicolumn{4}{c}{Active} & \multicolumn{2}{c}{Estimated} \\
Task        & SS        & RGB      & D        & EF       & MF       & LF       & D        & EF         
&           & SS        & RGB      & D        & EF       & MF       & LF       & D        & EF \\
\midrule
& \multicolumn{8}{c}{Training Conditions} && \multicolumn{8}{c}{New Town} \\
\midrule
Straight    &  $98.00\pm1.73$ & $96.33\pm1.53$ & $\underline{98.67\pm1.53}$ & $\underline{98.33\pm0.58}$ & $92.33\pm2.08$ & $\underline{\textbf{99.00}\pm0.00}$ & $92.33\pm1.15$ & $97.33\pm1.15$               &&   
              $100.00\pm0.00$ & $84.00\pm2.00$ & $94.33\pm0.58$ & $\underline{\textbf{96.33}\pm0.58}$ & $87.00\pm1.00$ & $77.00\pm0.00$ & $78.33\pm1.53$ & $71.67\pm2.08$ \\
One turn    & $100.00\pm0.00$ & $95.00\pm0.00$ & $92.00\pm0.00$ & $\underline{\textbf{99.00}\pm0.00}$ & $91.67\pm2.08$ & $90.33\pm0.58$ & $84.67\pm1.15$ & $96.33\pm1.53$                                                            &&    
               $96.67\pm0.58$ & $68.00\pm1.00$ & $74.33\pm2.52$ & $\underline{\textbf{79.00}\pm1.73}$ & $\underline{78.00\pm2.65}$ & $58.67\pm2.08$ & $46.33\pm1.15$ & $47.00\pm1.00$ \\
Navigation  &  $96.00\pm0.00$ & $89.00\pm2.00$ & $89.33\pm2.08$ & $\underline{92.67\pm1.15}$ & $90.67\pm1.15$ & $\underline{93.67\pm0.58}$ & $75.33\pm1.15$ & $\underline{\textbf{94.33}\pm0.58}$                           &&    
               $96.00\pm0.00$ & $59.67\pm3.06$ & $85.33\pm1.15$ & $\underline{\textbf{90.00}\pm2.00}$ & $80.67\pm0.58$ & $52.33\pm0.58$ & $45.67\pm3.06$ & $46.67\pm3.06$ \\
Nav.Dynamic &  $92.00\pm1.00$ & $84.00\pm2.00$ & $82.67\pm0.58$ & $\underline{89.33\pm0.58}$ & $78.33\pm2.89$ & $\underline{89.00\pm2.65}$ & $71.00\pm1.00$ & $\underline{\textbf{89.67}\pm1.15}$                           &&    
               $99.33\pm0.58$ & $54.33\pm3.79$ & $70.33\pm1.15$ & $\underline{\textbf{84.33}\pm2.52}$ & $73.67\pm2.52$ & $55.67\pm2.31$ & $44.33\pm2.52$ & $46.67\pm4.04$ \\
\midrule
& \multicolumn{8}{c}{New Weather} && \multicolumn{8}{c}{New Town \& Weather} \\
\midrule
Straight    & $100.00\pm0.00$ & $84.00\pm0.00$ & $\underline{\textbf{99.33}\pm1.15}$ & $96.00\pm2.00$ & $94.67\pm3.06$ & $96.00\pm0.00$ & $92.00\pm2.00$ & $84.67\pm1.15$                                                &&   
              $100.00\pm0.00$ & $84.67\pm1.15$ & $\underline{\textbf{97.33}\pm1.15}$ & $\underline{\textbf{97.33}\pm2.31}$ & $88.67\pm1.15$ & $\underline{\textbf{97.33}\pm1.15}$ & $78.00\pm0.00$ & $89.33\pm1.15$ \\
One turn    & $100.00\pm0.00$ & $76.67\pm4.16$ & $\underline{\textbf{94.67}\pm2.31}$ & $\underline{\textbf{94.67}\pm2.31}$ & $\underline{94.00\pm2.00}$ & $\underline{92.00\pm2.00}$ & $\underline{93.33\pm2.31}$ & $80.67\pm1.15$   &&    
               $96.00\pm0.00$ & $66.67\pm4.62$ & $72.67\pm1.15$ & $\underline{\textbf{82.67}\pm2.31}$ & $69.33\pm3.06$ & $67.33\pm2.31$ & $62.67\pm1.15$ & $64.00\pm3.46$ \\
Navigation  &  $95.33\pm1.15$ & $72.67\pm2.31$ & $89.33\pm1.15$ & $91.33\pm2.31$ & $90.67\pm3.06$ & $\underline{\textbf{96.00}\pm0.00}$ & $73.33\pm2.31$ & $80.67\pm5.03$                                                            &&    
               $96.00\pm0.00$ & $57.33\pm6.11$ & $84.00\pm3.46$ & $\underline{\textbf{92.67}\pm3.06}$ & $78.67\pm3.06$ & $72.67\pm1.15$ & $55.33\pm6.11$ & $60.67\pm2.31$ \\
Nav.Dynamic &  $92.67\pm1.15$ & $68.67\pm4.62$ & $\underline{90.00\pm2.00}$ & $86.00\pm4.00$ & $80.67\pm3.06$ & $\underline{\textbf{92.67}\pm3.06}$ & $76.67\pm4.16$ & $77.33\pm6.11$                                                &&    
               $98.00\pm2.00$ & $46.67\pm6.43$ & $69.33\pm2.31$ & $\underline{\textbf{94.00}\pm0.00}$ & $73.33\pm3.06$ & $73.33\pm2.31$ & $54.00\pm4.00$ & $49.33\pm3.06$ \\
\bottomrule
\end{tabular}
}
\end{table*}

\begin{table}
\caption{Succ. rate comparison with previous methods (see main text).}\label{tab:sota-comparison}
\centering
\resizebox{0.49\textwidth}{!}{
\begin{tabular}{@{}lccccccccccccccc@{}}
\toprule
Task        & MP & RL & CAL                      & CIRL                     & MT                       & Active EF                           && 
              MP & RL & CAL                      & CIRL                     & MT                       & Active EF \\
\midrule
            & \multicolumn{6}{c}{Training Conditions} && \multicolumn{6}{c}{New Town} \\
\midrule
Straight    & 98 & 89 & \underline{\textbf{100}} & 98                       & 98                       & $98.33\pm0.58$                      && 
              92 & 74 & 93                       & \underline{\textbf{100}} & \underline{\textbf{100}} & $96.33\pm0.58$ \\
One turn    & 82 & 34 & 97                       & 97                       & 87                       & $\underline{\textbf{99.00}\pm0.00}$ && 
              61 & 12 & \underline{\textbf{82}}  & 71                       & 81                       & $79.00\pm1.73$ \\
Navigation  & 80 & 14 & \underline{92}           & \underline{\textbf{93}}  & 81                       & $\underline{92.67\pm1.15}$          && 
              24 &  3 & 70                       & 53                       & 72                       & $\underline{\textbf{90.00}\pm2.00}$ \\
Nav.dynamic & 77 &  7 & 83                       & 82                       & 81                       & $\underline{\textbf{89.33}\pm0.58}$ && 
              24 &  2 & 64                       & 41                       & 53                       & $\underline{\textbf{84.33}\pm2.52}$ \\
\midrule
            & \multicolumn{6}{c}{New Weather} && \multicolumn{6}{c}{New Town \& Weather} \\
\midrule
Straight    & \underline{\textbf{100}} & 86 & \underline{\textbf{100}} & \underline{\textbf{100}} & \underline{\textbf{100}} & $96.00\pm2.00$             && 
              50                       & 68 & 94                       & \underline{\textbf{98}}  & \underline{96}           & $\underline{97.33\pm2.31}$ \\
One turn    & \underline{95}           & 16 & \underline{\textbf{96}}  & \underline{94}           & 88                       & $\underline{94.67\pm2.31}$          && 
              50                       & 20 & 72                       & \underline{82}           & \underline{82}           & $\underline{\textbf{82.67}\pm2.31}$ \\
Navigation  & \underline{\textbf{94}}  &  2 & 90                       & 86                       & 88                       & $\underline{91.33\pm2.31}$ && 
              47                       &  6 & 68                       & 68                       & 78                       & $\underline{\textbf{92.67}\pm3.06}$ \\
Nav.dynamic & \underline{\textbf{89}}  &  2 & 82                       & 80                       & 80                       & $\underline{86.00\pm4.00}$          && 
              44                       &  4 & 64                       & 62                       & 62                       & $\underline{\textbf{94.00}\pm0.00}$ \\
\bottomrule
\end{tabular}
}
\end{table}

\begin{table}
\caption{Infractions on dynamic navigation in new town \& weather. }\label{tab:CARLA-benchmark-detailed}
\centering
\resizebox{0.4\textwidth}{!}{
\begin{tabular}{@{}lccccccc@{}}
\bottomrule
Km per & Event & RGB & Active D  & Active EF \\
\midrule
\midrule
\multirow{2}{*}{Infraction}
                 & Sidewalk       & $0.86\pm0.10$  & $35.80\pm1.30$ & $16.76\pm5.54$ \\
                 & Opposite lane     & $0.73\pm0.04$  & $1.65\pm0.24$  & $3.29\pm1.96$ \\
\midrule
\multicolumn{2}{c}{Driven Km} (Perfect driving: 17.30 Km)    & $13.62\pm0.67$ & $35.80\pm1.30$ & $20.22\pm0.54$ \\
\bottomrule
\end{tabular}
}
\end{table}

During a training run we use Adam optimizer with 120 training samples per iteration (minibatch), an initial learning rate of 0.0002, decreased to the half each 50K iterations. Minibatches are balanced in terms of per $\actionbranch^\navcom$ branch samples. We set $\weights=(0.5, 0.45, 0.05)$ to weight the control signals (action) in the loss function. Action and speed losses are balanced by $\beta=0.95$. For selecting the best intermediate model of a training run, we do 500K iterations monitoring a validation performance measurement, $V_P$, each 100K iterations (thus, five times). The intermediate model with highest $V_P$ is selected as the resulting model of the training run. Since CIL models are trained from the scratch, variability is expected in their performance. Thus, for each type of model we perform five training runs, finally selecting the model with the highest $V_P$ among those resulting from the five training runs. 

Using Table~\ref{tab:t-v-t} as reference, we define $V_P$ to balance training-validation differences in terms of town and weather conditions. In particular, we use $V_P=0.25V_{w}+0.25V_{t}+0.50V_{wt}$; where $V_{w}$ is the success rate when validating in Town~1 and `soft rainy sunset' weather (not included in training data), $V_{t}$ is a success rate when validating in Town~2 (not included in training data) and `clear noon' weather (included in training data), and $V_{wt}$ stands for success rate when validating in Town~2 and `soft rainy sunset' (neither town, nor weather are part of the training data). Therefore, note that $V_P$ is a weighted success rate based on 75 episodes. 

\subsection{Training protocol}
\label{ssec:carla-training}

All CIL models in this paper rely on the same training protocol, partially following \cite{Codevilla:2018}. In all our CIL models original sensor channels (R/G/B/D) are trimmed to remove sky and very close areas (top and bottom part of the channels), and down-scaled to finally obtain channels of $200\times88$ pixel resolution. In our initial experiments, we found that traditional photometric and geometric recipes for data augmentation were not providing better driving models, thus, we do not use them. 

\subsection{Experimental results}
\label{ssec:carla-results}

We start the analysis of the experimental results by looking at Table \ref{tab:VV}, which is produced during training and selection of the best CIL networks. We focus first on RGB data as well as depth based on the post-processed CARLA depth ground truth, termed here as \emph{active} depth (\SSect{carla-dataset}) since its accuracy and covered depth range is characteristic of active sensors ({\eg} LiDAR). We see that the best (among five training runs) validation performance $V_P$ is $48\%$ when using RGB data only. So we will use the corresponding CIL model as RGB-based driver in the following experiments. Analogously, for the case of using only active depth (D), the best CIL reports a performance of $74\%$. The best performances for early fusion (EF), mid fusion (MF), and late fusion (LF) are $91\%, 74\%$ and $67\%$, respectively. Again, we take the corresponding CIL models as drivers for the following experiments.   

Table \ref{tab:CARLA-benchmark} reports the performance of the selected models according to the original CARLA benchmark. We have included a model trained on perfect semantic segmentation (SS) according to the five classes considered here for self-driving (see \Fig{rgb-gt-monocular}). Thus, we consider this model as an upper bound. Indeed, its performance is most of the times $\geq96$, reaching $100$ several times. This also confirms that the CIL model is able to drive properly in CARLA conditions provided there is a proper input. We can see that, indeed, active depth is a powerful information for end-to-end driving by itself, clearly outperforming RGB in non-training conditions. However, in most of the cases RGBD outperforms the use of only RGB or only D. The most clear case is for new town and weather with dynamic objects, {\ie} for the most challenging conditions, where RGB alone reaches a success rate of $46.67\pm6.43$, D alone $69.33\pm2.31$, but together the success rate is $94.00\pm0.00$ for early fusion.
For a new town (irrespective of the weather conditions) early fusion clearly outperforms mid and late fusion.
In any case, it is clear that multimodality improves CIL performance with respect to a single modality, which is the main question we wanted to answer in this paper. 

In order to further analyse the goodness of multimodality, we compare it to previous single-modality methods (see \Sect{rw}). Not all the corresponding papers provide details about the training methodology or training datasets; thus, this comparison is solely based on the reported performances on the original CARLA benchmark and must be taken only as an additional reference about the goodness of multimodality. Early fusion, is the smaller CNN architecture in terms of weights, thus, we are going to focus on it for this comparison. Table \ref{tab:sota-comparison} shows the results. MP and RL stand for modular perception and reinforcement learning, respectively. The reported results are reproduced from \cite{Dosovitskiy:2017}. CAL stands for conditional affordance learning and the results are reproduced from \cite{Sauer:2018}. CIRL stands for controllable imitative reinforcement learning and the results are reproduced from \cite{Liang:2018}. Finally, MT stands for multi-task learning, and the results are reproduced from \cite{LiMotoyoshi:2018}. We see how, in presence of dynamic traffic participants, 
the RGBD early fusion (with active depth) is the model with higher success rate on the original CARLA benchmark. On the other hand, such an early fusion approach can be combined with CAL or CIRL methods, they are totally compatible. We think that this comparison with previous works reinforces the idea that multimodality can help end-to-end driving.

Once it is clear that multimodality is beneficial for end-to-end driving, we can raise the question of whether monocular depth estimation \cite{Godard:2017, Luo:2018, Xu:2018, Gurram:2018} can be as effective as depth coming from active sensors in this context. In the former case, it would consists on a multisensory multimodal approach, while the later case would correspond to a single-sensor multimodal approach since both RGB and depth come from the same camera sensor (depth is estimated from RGB). In order to carry out a proof-of-concept, we use our own monocular depth estimation model \cite{Gurram:2018} (it was state-of-the-art at the moment of its publication) fine-tuned on CARLA training dataset. More specifically, the dataset used for training the multimodal CIL models is also used to fine-tune our monocular depth estimation model, {\ie} using the post-processed depth channels and corresponding RGB images. During training, we monitor the regression loss until it is stable, we do not stop training based on the performance on validation episodes. Figure \ref{fig:rgb-gt-monocular} shows and example of monocular depth estimation.

Analogously to the experiments shown so far, we train a CIL model based on the estimated depth as well as on the corresponding multimodal (RGBD) fusion. In order to reduce the burden of experiments, we use early fusion since it is the best performing for the active depth case. The training performances for model selection can be seen in Table \ref{tab:VV}. We use the CIL models of $V_P$ $59\%$ and $51\%$, respectively. In validation terms, such performances are already clearly worse than the analogous based on active depth. Table \ref{tab:CARLA-benchmark} shows the results on the original CARLA benchmark. Indeed, these are worse than using active depth, however, still when remaining in the 
training conditions monocular-based EF outperforms depth and RGB alone, and in fact shows similar performance as active depth.
This is not the case when we change from training conditions since monocular depth estimation itself does not perform equally well in this case, and so happens to EF. 
However, we think that this single-sensor multimodal setting is really worth to pursuit. Moreover, although it is out of the scope of this paper, we think that performing end-to-end driving may be a good protocol for evaluating depth estimation models beyond the static metrics currently used, which are agnostic to the task in which depth estimation is going to be used. Note that even for evaluating the driving performance of end-to-end driving models in itself, it has been shown that relying only on static evaluations may be misleading \cite{Codevilla:2018b, Bewley:2018}.

Finally, for the RGB, Active D and EF models, we assess additional infractions for new town and weather conditions with dynamic objects. Table \ref{tab:CARLA-benchmark-detailed} shows the driven Km per infraction of each model. Note that not all such infractions imply an accident stopping the AV. For instance, the AV can run into an opposite lane a bit without crashing with other vehicles. As a reference, we also show the amount of driven Km in which these measurements are based. All models are supposed to complete the same testing routes ({\ie}, same total Km), termed as \emph{perfect driving} in Table \ref{tab:CARLA-benchmark-detailed}. However, if a model fails to follow the right path at an intersection the route would be recomputed, thus, it will need more Km to reach the destination. On the contrary, if it fails to complete the routes, the driven Km will be lower. We see that RGB and Active EF models are not far, but Active D failed too much at taking the right path at intersections. RGB performs the worst in all metrics. The Active D model does not run over the sidewalk and uses the curbside as a cue that also helps on lane keeping except at intersections. Active D shows a good equilibrium between RGB and Active depth single-modality models.

\section{Conclusion}
\label{sec:conclusion}

In this paper, we compare single- and multimodal perception data for end-to-end driving. As multimodal perception data we focus on RGB and depth, since they are usually available in autonomous vehicles through the presence of cameras and active sensors such as LiDAR. As end-to-end driving model we use branched conditional imitation learning (CIL). Relying on a well-established simulation environment, CARLA, we assess the driving performance of single-modal (RGB, depth) CIL models, as well as multimodal CIL models according to early, mid, and late fusion paradigms. In all cases, the depth information available in CARLA is post-processed to obtain a more realistic range of distances and depth accuracy. This depth is also used to train a depth estimation model so that the experiments  cover multimodality not only based on a multisensory setting (RGB and active depth) but also based on a single-sensor setting (RGB and estimated depth). Overall, the experiments clearly allow us to conclude that multimodality (RGBD) is indeed a beneficial approach for end-to-end driving. In fact, we plan to follow this line of work in the near future, focusing on the single-sensor setting since better estimation models are required in order to compete with the multisensory setting. In addition, we also plan to consider other sources of multi-modality usually available in modern vehicles, such as GNSS information which, even usually being noisy, eventually can complement direct scene sensing.


\bibliographystyle{IEEEtran}
\bibliography{references}

\ifCLASSOPTIONcaptionsoff
  \newpage
\fi

\end{document}